\documentclass[sigconf]{acmart}
\usepackage{multirow}
\usepackage{arydshln}
\usepackage{longtable}
\usepackage{multirow}
\usepackage{hhline}
\usepackage{tabularray}
\usepackage{balance}
\usepackage[font={small}]{caption, subfig}
\usepackage{orcidlink}

\AtBeginDocument{%
  }

\setcopyright{acmlicensed}
\copyrightyear{2018}
\acmYear{2018}
\acmDOI{XXXXXXX.XXXXXXX}

\acmConference[Conference acronym 'XX]{Make sure to enter the correct
  conference title from your rights confirmation emai}{June 03--05,
  2018}{Woodstock, NY}
\acmISBN{978-1-4503-XXXX-X/18/06}

\newtheorem{problem}{Problem}

\newtheorem{task}{Task}

\makeatletter
\def\subsubsection{\@startsection{subsubsection}{3}%
  \z@{.5\linespacing\@plus.7\linespacing}{.1\linespacing}%
  {\normalfont\itshape}}
\makeatother

\copyrightyear{2024}
\acmYear{2024}
\setcopyright{rightsretained}
\acmConference[CIKM '24]{Proceedings of the 33rd ACM International
Conference on Information and Knowledge Management}{October 21--25,
2024}{Boise, ID, USA}
\acmBooktitle{Proceedings of the 33rd ACM International Conference on
Information and Knowledge Management (CIKM '24), October 21--25, 2024,
Boise, ID, USA}
\acmDOI{10.1145/3627673.3680044}
\acmISBN{979-8-4007-0436-9/24/10}

\begin{document}

\title{\textsc{LAPIS}: Language Model-Augmented Police Investigation System}





\author{Heedou Kim\orcidlink{0009-0009-0140-1264}}
\orcid{0009-0009-0140-1264}
\affiliation{%
  \institution{Korea University}
  \city{Seoul}
  \country{South Korea}
}
\affiliation{%
  \institution{Police Science Institute}
  \city{Asan}
  \country{South Korea}
}
\email{heedou123@korea.ac.kr}

\author{Dain Kim\orcidlink{0009-0000-1984-8982}}
\orcid{0009-0000-1984-8982}
\affiliation{%
  \institution{Korea University}
  \city{Seoul}
  \country{South Korea}
}
\email{dain-kim@korea.ac.kr}

\author{Jiwoo Lee\orcidlink{0009-0008-1787-3664}}
\orcid{0009-0008-1787-3664}
\affiliation{%
  \institution{Korea University}
  \city{Seoul}
  \country{South Korea}
}
\email{hijiwoo7@korea.ac.kr}

\author{Chanwoong Yoon\orcidlink{0009-0004-6038-8702}}
\orcid{0009-0004-6038-8702}
\affiliation{%
  \institution{Korea University}
  \city{Seoul}
  \country{South Korea}
}
\email{cwyoon99@korea.ac.kr}

\author{Donghee Choi\orcidlink{0000-0002-8857-9680}}
\orcid{0000-0002-8857-9680}
\affiliation{%
  \institution{Imperial College London}
  \city{London}
  \country{United Kingdom}
}
\email{donghee.choi@imperial.ac.uk}

\author{Mogan Gim$^1$\orcidlink{0000-0002-6458-7723}}
\authornotemark[2]
\orcid{0000-0002-6458-7723}
\affiliation{%
  \institution{Hankuk University of Foreign Studies}
  \city{Yongin}
  \country{South Korea}
}
\email{akim@korea.ac.kr}

\author{Jaewoo Kang\orcidlink{0000-0001-6798-9106}}
\authornotemark[2]
\orcid{0000-0001-6798-9106}
\affiliation{%
  \institution{Korea University}
  \city{Seoul}
  \country{South Korea}
}
\email{kangj@korea.ac.kr}
\thanks{$^\dagger$ Corresponding author}
\thanks{$^1$ This work was done while the author was a postdoctoral researcher at Korea University.}
\renewcommand{\shortauthors}{Kim, et al}



\begin{abstract}
Crime situations are race against time. An AI-assisted criminal investigation system, providing prompt but precise legal counsel is in need for police officers. We introduce LAPIS (Language Model Augmented Police Investigation System), an automated system that assists police officers to perform rational and legal investigative actions. We constructed a finetuning dataset and retrieval knowledgebase specialized in crime investigation legal reasoning task. We extended the dataset's quality by incorporating manual curation efforts done by a group of domain experts. We then finetuned the pretrained weights of a smaller Korean language model to the newly constructed dataset and integrated it with the crime investigation knowledgebase retrieval approach. Experimental results show LAPIS' potential in providing reliable legal guidance for police officers, even better than the proprietary GPT-4 model. Qualitative analysis on the rationales generated by LAPIS demonstrate the model's reasoning ability to leverage the premises and derive legally correct conclusions.
\end{abstract}



\received{20 February 2007}
\received[revised]{12 March 2009}
\received[accepted]{5 June 2009}



\maketitle

\begin{figure}[h]
    \scalebox{0.8}
    {
    \centering
    \includegraphics[width=0.48\textwidth]{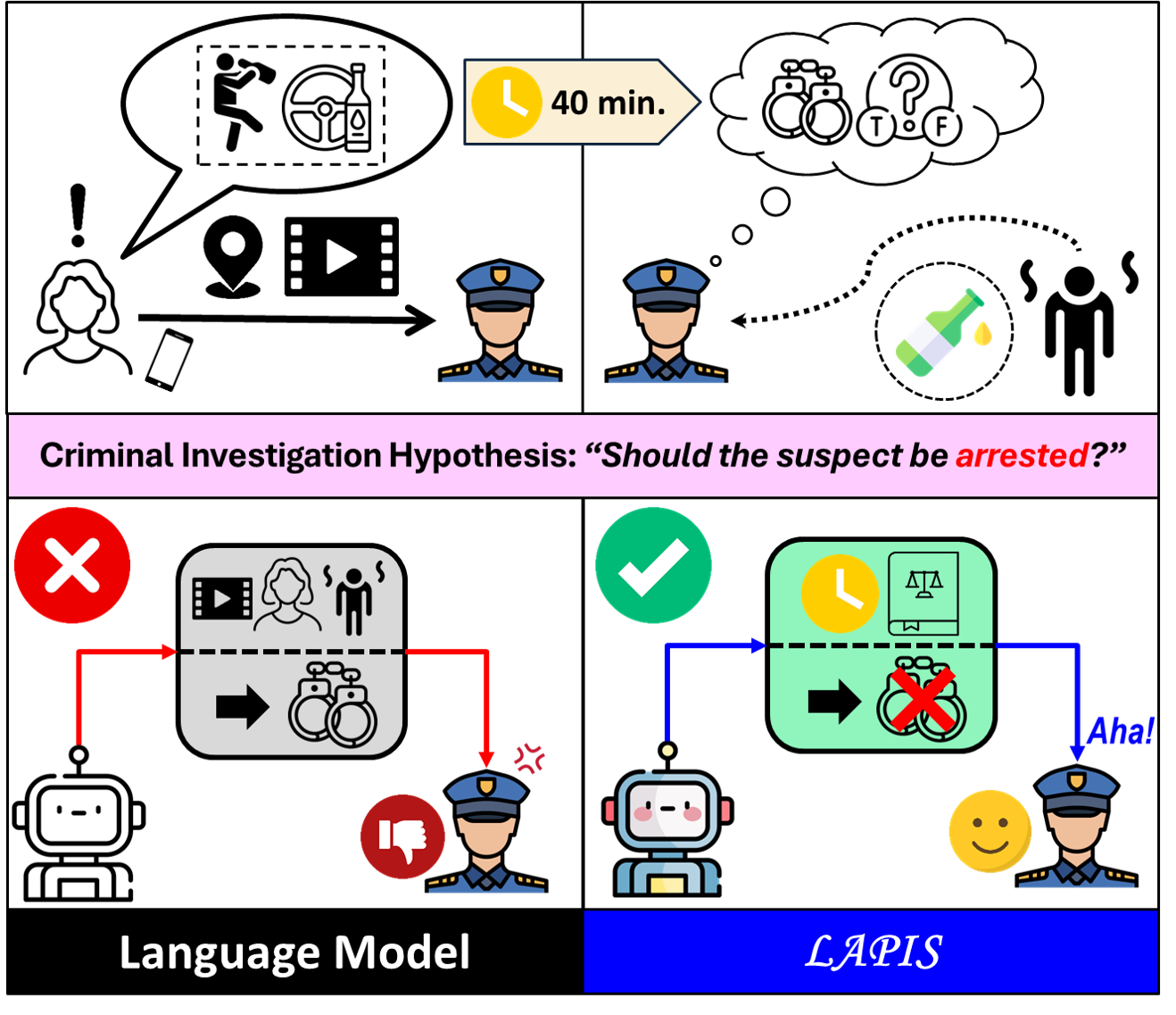}
	}
    \caption{
	In a crime investigation scenario, different outcomes emerged based on hypothesis assessment, with a non-domain specific LLM struggling to answer accurately, whereas LAPIS provided a precise response with a relevant premise.
	}
	\label{fig:overview}
\end{figure}

\section{Introduction}

Police officers are gradually overburdened with increasing numbers of diversified crimes which may lead to shorthanded crime investigation. Particularly in South Korea, the incidence of cyber crimes has more than doubled over the eight-year period from 2014 to 2022, increasing from 110,109 cases to 230,355 cases~\cite{KoreaNationalPoliceAgency2022}. At the same time, the average number of days required to process a reported crime per police officer has steadily increased from 50.4 days in 2019 to 67.7 days in 2022~\cite{KoreaNationalPoliceAgency2022}. Moreover, considering increasingly complex nature of recent crimes in our evolving society, the most intellectually challenging and time-consuming aspect for police officers is determining whether their actions are procedurally and legally justifiable~\cite{bottoms2012beyond}. This leads to the necessity of using and developing automated approaches in AI-assisted law enforcement such as crime prediction and investigation~\cite{hunter2019framework}. 

Crime investigation is the process within the criminal justice system of proving the occurrence of a criminal act when there is a possibility that certain behavior or its outcomes may constitute a crime~\cite{roberts2012law}. Given the current context of crime investigation such as collected evidence or analysis results, police officers perform \textit{investigative actions} based on their \textit{legal reasoning} skills~\cite{clark2012covert}. \textit{Legal reasoning} commonly involves legal assessment of \textit{hypotheses} based on its supportive \textit{premises}~\cite{yu2022legal}. These \textit{premises} formulated based on crime investigation knowledge constitute a logical \textit{rationale} that explains the \textit{hypothesis} assessment. This concept is crucial as it guides crime investigators within the legal boundaries stated by the criminal law system. 

However, one of the challenges in developing AI-assisted law enforcement applications is the difficulty of incorporating the intangible concept of legal reasoning~\cite{choi2023ai, chalkidis2023chatgpt}. Moreover, the important traits in crime investigation are swift and accurate decision-making and responsive action~\cite{o1956criminal}. To address, we propose to borrow the foundational ideas and methods from pre-existing domain-specific application systems of other tasks such as clinical event prediction and decision-making~\cite{xu2024survey,kim2024meerkat,tian2024chatgptbio,thirunavukarasu2023lm_medicine}. The key pillars of these systems involve the use of large language models (LLM) and their injection of expert-level knowledge. 

Locally trained small language models (SLM) have gained attraction in developing domain-specific application systems~\cite{ho2022large, magister2022teaching}. While utilizing proprietary LLMs via its API (e.g GPT-4) could potentially benefit the development of a crime investigation system, there are several weaknesses in terms of their practicability. First, such LLMs cannot be further trained with continuously updated laws and confidential police investigation data. Also, there is a security risk where sensitive information such as ongoing crime cases could potentially be exposed through police officers' API usage~\cite{zhou2023ethical, liu2023trustworthy, yan2024protecting}. Lastly, public LLMs are susceptible to ethical complications such as privacy leakage or the misuse of criminal records used for training~\cite{falade2023decoding, lukas2023analyzing, kim2024propile}. This further justifies the need of employing locally trained SLMs when developing an AI-assisted police investigation system.

Such local SLMs can be obtained by fine-tuning their pretrained weights utilizing a well-designed instruction dataset~\cite{xu2024survey, kang2024knowledge}. Our work emphasizes addressing the critical elements of legal reasoning through the injection of crime investigation knowledge. A fine-tuned SLM lacking these elements, when embedded in a crime investigation system, can generate incorrect legal assessments, potentially leading police officers to perform misguided investigative actions~\cite{vestby2021machine, dahl2024large}. To address this, we propose to employ retrieval-augmented generation (RAG) from a domain-specific knowledgbase in both finetuning dataset construction and the SLM's functionality. Moreover, we incorporated manual curation efforts performed by a group of domain experts to further the quality of our dataset.

As illustrated in Figure~\ref{fig:overview}, suppose a police officer received an emergency call where the suspect was witnessed driving under the influence. Having encountered the suspect after 40 minutes passed, the police officer felt a strong scent of alcohol from them. Due to these circumstances, a \textit{hypothesis} rises whether the police officer can arrest the suspect, who has allegedly committed a crime of driving under the influence, or not. Given this \textit{hypothesis} as input prompt, an LLM-based agent would respond that the suspect should be indeed arrested for their alleged crime as there are substantial evidence. However, the criminal law states that the officer \textbf{cannot} exercise their lawful rights to arrest the suspect, as arresting a driver who has been at the roadside for more than 40 minutes after the initial report cannot be deemed a lawful execution of official duty. This highlights the importance of equipping LLMs with expert-level CI knowledge in order to elude wrongful lawful executions such as misarrests. 

In this work, we present a \textbf{L}anguage Model-\textbf{A}ugmented \textbf{P}olice \textbf{I}nvestigation \textbf{S}ystem (LAPIS). LAPIS guides the police officer's decision-making on investigative actions through generation of accurate assessments and their rationale given hypothesis questions, empowered by legal reasoning. The building blocks that address the design principles of developing a crime investigation system are our novel task formulation of \textbf{C}rime \textbf{I}nvestigation, and our newly constructed \textbf{C}rime \textbf{I}nvestigation \textbf{L}egal \textbf{R}easoning (\textbf{CIRL}) dataset and \textbf{C}rime \textbf{I}nvestigation \textbf{K}nowledge\textbf{B}ase (\textbf{CIKB}) which are for SLM finetuning and RAG purposes respectively. To the best of our knowledge, our work firstly attempts to develop a system specialized in crime investigation task.

Our experiment design explores the effects of finetuning various pretrained SLMs, providing each training instance with different types of supportive rationales (i.e expert-curated, GPT-generated) and utilizing domain-specific RAG module. Furthermore, we deployed LAPIS in a crime investigation scenario and evaluated not only its hypothesis assessment but also generated supportive rationales. Both the experimental and simulation results conclude that LAPIS can potentially benefit crime investigation in police agencies. Code and data is available in our github repository~\footnote{https://github.com/dmis-lab/LAPIS}.

The main contributions of our work are the following,
\begin{itemize}
    \item We formulated a downstream task for LAPIS called crime investigation which involves legal reasoning and knowledge retrieval.
    \item We constructed an expert-curated finetuning dataset and knowledgebase to effectively inject crime investigation knowledge in LAPIS' SLM. 
    \item Both quantitative and qualitative results demonstrate LAPIS' potential in benefiting police officers and law enforcement agencies.
\end{itemize}

\section{Related Work}
\subsection{AI in Law Enforcement}
The applicability of AI in law enforcement has recently gained recognition especially by fields of forensic science and criminal investigation~\cite{bag2024using,maneli20223d,richmond2020ai}. Moreover, law enforcement agencies have recently started to incorporate AI-driven automated systems for aiding police officers, crime investigators and forensics experts~\cite{jadhav2020artificial}. Particularly, researchers have explored various ways to utilize AI-related methodologies in crime investigation (CI) domain. \citeauthor{johnsen2019impact} emphasized the importance of following text preprocessing requirements when utilizing Open-source Intelligence (OSINT) to preemptively deal with cyber threats~\cite{johnsen2019impact}. \citeauthor{schiliro2021novel} devised a novel cognitive computing approach using convolutional neural networks to develop a cognitive assistant for police investigators while \citeauthor{bunnin2021bayesian} introduced a novel archetype of Bayesian hierarchical models for inferring progression of criminal attacks~\cite{schiliro2021novel,bunnin2021bayesian}. \citeauthor{hepenstal2021developing} proposed a novel design approach for developing conversational agents and \citeauthor{kim2022named} implemented a deep learning-based named recognition model, both specialized in CI~\cite{hepenstal2021developing,kim2022named}. Motivated by these works, our work employs expert-curated dataset construction techniques and conventional LLM-related methodologies to build a system that provides precisive legal guidance for police investigators.

\subsection{LLMs and Legal Reasoning}
Legal practice is one of the fundamental aspects of crime investigation. Recently, LLMs have been trained with massive parameter size, enabling them to exhibit impressive capabilities in numerous NLP tasks through zero-shot or few-shot prompting learning~\cite{brown2020language,chowdhery2023palm}. Advancements in LLMs also have a significant impact on the utilization of AI in the legal field~\cite{nguyen2023beyond}. \citeauthor{chalkidis2023chatgpt} evaluated LLMs on LexGLUE benchmark dataset~\cite{chalkidis2021LexGLUE} in a zero-shot manner, and~\citeauthor{katz2023gpt} evaluated GPT-4 on bar exams by using instruction-following formats. Additionally, there are on-going studies that verify the effectiveness of legal prompting by inducing legal reasoning skills to LLMs~\cite{trautmann2022legal}. \citeauthor{jiang2023legal} introduced legal syllogism prompting to teach the general legal thought process to LLM for legal judgment prediction. \citeauthor{yu2022legal} added an IRAC explanation to instructions, which provides a skill for thinking like a lawyer, and guided the process of legal reasoning through a few-shot demonstration on COLIEE's legal entailment task. While these studies explore LLM's general legal reasoning capabilities mostly tailored for law practitioners, our work focuses on developing a LLM-powered crime investigation system which requires reasoning skills specialized in criminal law and investigation principles.
\begin{figure*}[h]
    \scalebox{0.8}
    {
    \centering
    \includegraphics[width=0.99\textwidth]{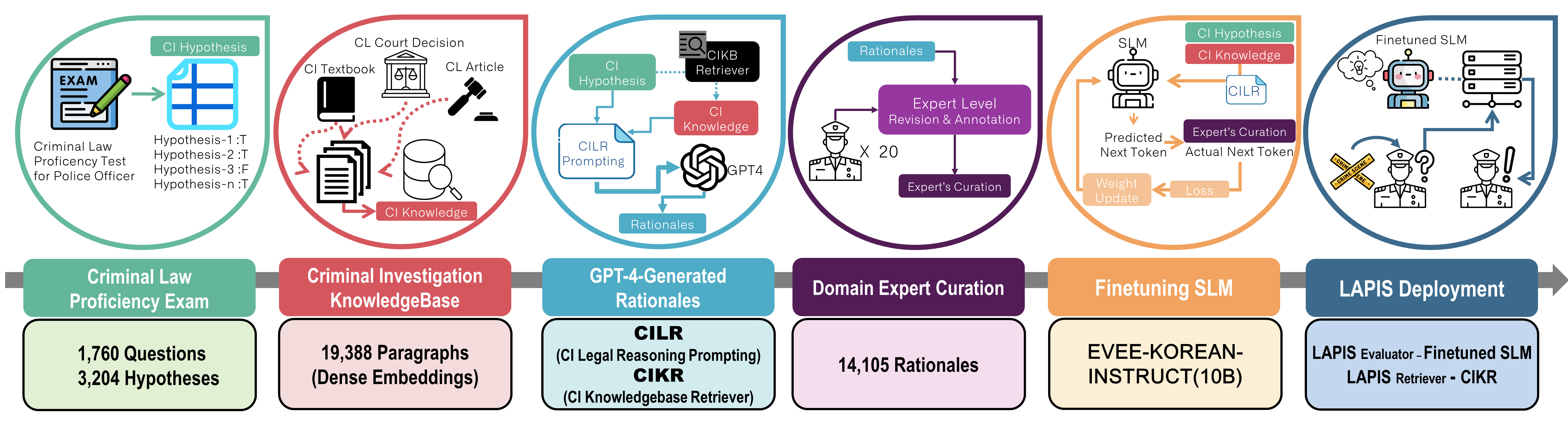}
	}
    \caption{Illustrative workflow of LAPIS development process.}
	\label{fig:workflow}
\end{figure*}

\section{Methods}
\subsection{Task Formulation}
\begin{problem}[Crime Investigation Task]
\label{problem:ci_hypothesis}
We define the objective of \textbf{Crime Investigation Task} as finite steps of performing \textbf{Legal Reasoning} and \textbf{Investigative Action} with the objective of proving or disapproving a criminal allegation or suspicion. Each step involves LAPIS' legal reasoning task which guides the police officer's decision making on the appropriate investigative action to take. 
\end{problem}

To solve the above problem, we formulate the following two tasks performed by the components of LAPIS, evaluator ($f$) and retriever ($g$).

\begin{task}[Legal Reasoning]
\label{task:task_legal_reasoning}
Given an input \textbf{prompt} $\mathbf{X}$ questioning whether the \textit{hypothesis} $h$ formulated under the current crime investigation context $\mathcal{C}$ is true or false, LAPIS evaluator $f$ generates a \textbf{response} $\mathbf{y}$ consisting a pair of True-False \textbf{assessment} $y_a\in[0,1]$ and its corresponding \textbf{rationale} $y_r$. As a result, this task is modeled by $\mathbf{y} = f(\mathbf{X})$ where $\mathbf{X}=\left\{h, \mathcal{C}\right\}$ and $\mathbf{y}=\left\{y_a, y_r\right\}$.
\end{task}

\begin{task}[Knowledge Retrieval]
\label{task:task_knowledge_retrieval}
Given an input \textbf{prompt} $\mathbf{X}$, LAPIS retriever $g$ retrieves the most relevant \textbf{premises} $\mathbb{P}$ from crime investigation \textbf{knowledgebase} $\mathcal{K}$ and exploits them in generating its \textbf{response} $\mathbf{y}$. The enhanced version of LAPIS' legal reasoning task is thus modeled by $\mathbf{y} = f(\mathbf{X}, \mathbb{P})$ where $\mathbb{P} = g(\mathbf{X})$. 
\end{task}

Figure~\ref{fig:workflow} shows the development process of LAPIS starting from task formulation to deployment. We elaborate the following details of developing LAPIS in this section which mainly involve dataset construction, knowledgbase creation, domain expert curation and SLM finetuning.

\subsection{Dataset Construction}

\begin{table*}
\centering
\scalebox{0.6}
{
\begin{tabular}{|l|c|c|c|c|c|c|c|c|c|c|c|c|c|c|} 
\hline
                                & \textbf{Original Dataset} & \textbf{Initial Dataset}                & \multicolumn{8}{c|}{\textbf{Training Set}}                                                                                                                                                                                                                                                                                                                                                                                                                                                                                                                                                                                                                                                                                                                                                         & \multicolumn{3}{c|}{\textbf{Development Set}}                                                                                                                                                                                                  & \textbf{Test Set}                        \\ 
\hline
\textbf{Year Period}            & 2013-2023                 & 2013-2023                               & \multicolumn{8}{c|}{2013-2019}                                                                                                                                                                                                                                                                                                                                                                                                                                                                                                                                                                                                                                                                                                                                                                     & \multicolumn{3}{c|}{2020}                                                                                                                                                                                                                      & 2021-2023                                \\ 
\hline
\textbf{Data Instance Type}     & \textit{Exam Questions}   & $\mathcal{D}_i$ $(\mathcal{C}, h, y_a)$ & $\mathcal{D}_t$ $(\mathcal{C}, h, y_a)$ & \multicolumn{7}{c|}{$\mathcal{D}^{+}_t$$(\mathcal{C}, h, \mathbb{P}, y_a, y_r)$}                                                                                                                                                                                                                                                                                                                                                                                                                                                                                                                                                                                                                                                                         & $\mathcal{D}_d$ $(\mathcal{C}, h, y_a)$ & \multicolumn{2}{c|}{$\mathcal{D}^{+}_d$ $(\mathcal{C}, h, \mathbb{P}, y_a, y_r)$}                                                                                                                    & $\mathcal{D}_s$ $(\mathcal{C}, h, y_a)$  \\ 
\hline
\textbf{Rationale Type}         & -                         & -                                       & -                                       & \multicolumn{1}{l|}{\begin{tabular}[c]{@{}l@{}}\textbf{GPT4-}\\\textbf{VP-}\\\textbf{ZS}\end{tabular}} & \multicolumn{1}{l|}{\begin{tabular}[c]{@{}l@{}}\textbf{GPT4-}\\\textbf{IRAC-}\\\textbf{ZS}\end{tabular}} & \multicolumn{1}{l|}{\begin{tabular}[c]{@{}l@{}}\textbf{GPT4-}\\\textbf{IRAC-}\\\textbf{1S}\end{tabular}} & \multicolumn{1}{l|}{\begin{tabular}[c]{@{}l@{}}\textbf{GPT4-}\\\textbf{CILR-}\\\textbf{ZS}\end{tabular}} & \multicolumn{1}{l|}{\begin{tabular}[c]{@{}l@{}}\textbf{GPT4-}\\\textbf{CILR-}\\\textbf{1S}\end{tabular}} & \multicolumn{1}{l|}{\begin{tabular}[c]{@{}l@{}}\textbf{GPT4-}\\\textbf{CILR-}\\\textbf{3S}\end{tabular}} & \multicolumn{1}{l|}{\begin{tabular}[c]{@{}l@{}}\textbf{DEXP-}\\\textbf{ANN}\end{tabular}} & -                                       & \multicolumn{1}{l|}{\begin{tabular}[c]{@{}l@{}}\textbf{GPT4-}\\\textbf{CILR-}\\\textbf{3S}\end{tabular}} & \multicolumn{1}{l|}{\begin{tabular}[c]{@{}l@{}}\textbf{DEXP-}\\\textbf{ANN}\end{tabular}} & -                                        \\ 
\hline
\textbf{Criminal Law}           & 440                       & 1,273                                    & 811                                     & 474                                                                                                    & 501                                                                                                     & 551                                                                                                      & 540                                                                                                      & 578                                                                                                      & 597                                                                                                      & 201                                                                                       & 120                                     & 97                                                                                                       & 27                                                                                        & 342                                      \\
\textbf{Criminal Procedure Law} & 440                       & 1,318                                    & 855                                     & 545                                                                                                    & 587                                                                                                     & 628                                                                                                      & 620                                                                                                      & 659                                                                                                      & 657                                                                                                      & 182                                                                                       & 124                                     & 99                                                                                                       & 27                                                                                        & 339                                      \\
\textbf{Crime Investigation}    & 880                       & 2,304                                    & 1,538                                    & 933                                                                                                    & 1,025                                                                                                    & 1,078                                                                                                     & 1,112                                                                                                     & 1,145                                                                                                     & 1,172                                                                                                     & 320                                                                                       & 218                                     & 171                                                                                                      & 44                                                                                        & 548                                      \\ 
\hline
\textbf{All Subject Areas}      & 1,760                      & 4,895                                    & 3,204                                    & 1,952                                                                                                   & 2,113                                                                                                    & 2,257                                                                                                     & 2,272                                                                                                     & 2,382                                                                                                     & 2,426                                                                                                     & 703                                                                                       & 462                                     & 367                                                                                                      & 98                                                                                        & 1,229                                     \\ 
\hline
\textbf{All Data Instances}     & -                         & 4,895                                    & 3,204                                    & \multicolumn{7}{c|}{\textbf{14,105}}                                                                                                                                                                                                                                                                                                                                                                                                                                                                                                                                                                                                                                                                                                                      & 462                                     & \multicolumn{2}{c|}{\textbf{465}}                                                                                                                                                                    & \textbf{1,229}                            \\
\hline
\end{tabular}
}
\caption{Detailed statistics of the Crime Investigation Dataset. The numbers of data instances containing input prompt texts of current crime investigation context ($\mathcal{C}$), hypothesis ($h$) and premises retrieved from the CI knowledgebase ($\mathbb{P}\subset\mathcal{K}$) and output response texts of True-False assessment ($y_a$), supportive rationale ($y_r$) are 14,105 and 465 for the training ($\mathcal{D}^{+}_t$) and development ($\mathcal{D}^{+}_d$) set respectively. The test set $\mathcal{D}_s$ solely used for quantitative evaluation purposes contains data instances only comprised of ($\mathcal{C},h,y_a$). \textbf{DEXP-ANN} refers to rationales manually annotated by domain experts.}
\label{tab:lapis_dataset}
\end{table*}

\subsubsection{Criminal Law Proficiency Exam}\hfill\\
The annual Criminal Law Proficiency Exam organized by the Korean National Police Agency evaluates the qualifications of police officers' lawful duty to conduct crime investigations. As this process requires swift and precise decision-making skills, examinees are required to read long texts implicating crime investigation context and assess each of its possible legal hypotheses in a limited amount of time. The examination consists of three subject areas which are Criminal Law, Criminal Procedure law and Crime Investigation. The first two subjects mostly comprise questions testing the examinee's knowledge in criminal law related to investigative procedures while the last subject covers questions related to the theoretical concepts and real-world protocols of crime investigation itself.

We collected the Korean-written exam questions from the three subjects ranging an yearly period from 2013 to 2023, with the total number being 1,760. Each subject contains multiple-choice questions where most of them explicitly state crime investigation context. To strictly evaluate LAPIS' hypothesis assessment ability, we partitioned the dataset on an annual basis into train ($\mathcal{D}_t$, 2013-2020), development ($\mathcal{D}_d$, 2020) and test ($\mathcal{D}_s$, 2021-2023). Inspired by the COLIEE dataset~\cite{yu2022legal}, we first converted all the given multiple-choice options in the exam questions into hypotheses answerable as true or false ($h$, $y_a$). As a result, our initial Crime Investigation Legal Reasoning dataset contains 4,895 data instances ($\mathcal{D}_i$ ($\mathcal{C}$, $h$, $y_a$)).

\subsubsection{Crime Investigation Knowledgebase}\hfill\\
Accurately answering hypothesis-based questions of crime investigation area with simple True or False responses does not fully suffice the qualifications of being a police investigator. Each investigative action performed by the police officer should adhere to the investigative protocols and requires concretely supportive rationale based on officially documented criminal law articles and previous court rulings~\cite{chaiken1977criminal,stelfox2013criminal,fahsing2016making}.

Based on this motivation, we created a Crime Investigation Knowledgebase (\textbf{CIKB}) containing documents from multiple data sources which are Criminal Investigation textbooks, up-to-date Criminal Law Articles and Criminal Court Rulings. The Criminal Investigation textbooks cover theories and practical methodologies such as behavioral analysis and interrogation~\cite{KoreanCriminalInvestigation2023}. The Criminal Law Articles collected from the Korean Law Information Center provide the fundamental legal guidelines for specifying criminal acts along with each of its degrees and forms of punishment~\cite{KoreanLawInformationCenter2023}. The Criminal Court Rulings also collected from the Korean Law Information Center, are actual judgements issued by the Supreme Court of Korea which contain comprehensive information on actual criminal cases~\cite{KoreanLawInformationCenter2023}. 

As a result, our newly created \textbf{CIKB} $\mathcal{K}$ contains 19,388 dense embeddings of 687 paragraphs (223,224 tokens) extracted from Criminal Investigation textbooks, 3,304 paragraphs (1,442,869 tokens) from Criminal Law Articles and 15,397 paragraphs (3,232,570 tokens) from Criminal Court Rulings. We generated these dense embeddings using a pretrained GPT3.5

\subsubsection{GPT-4-Generated Rationales}\hfill\\
Given the data instances ($\mathcal{C}$, $h$, $y_a$) in the training and development parition of our initial dataset $\mathcal{D}_i$, we prompted the GPT-4 to generate responses consisting its predicted true-false assessment and rationale. These GPT-generated rationales $y_r$ are included in the output part of each data instance which guides the SLM not only make correct true-false hypothesis assessment but also provide better explainability to police officers. 

When prompting the GPT-4, we employed two approaches which are utilizing our devised Crime Investigation Legal Reasoning (\textbf{CILR}) prompting method and a RAG method that exploits our created \textbf{CIKB} denoted as Crime Investigation Knowledge Retrieval (\textbf{CIKB}). The \textbf{CILR} prompting method is a modified version of the Issue-Rule-Application-Conclusion (IRAC) prompting method, previously used in legal entailment tasks which are distant from actual crime investigation scenarios~\cite{yu2022legal}. The \textbf{CIKR} method performs a vector similarity search on the \textbf{CIKB} given the crime investigation context ($\mathcal{C}$) and hypothesis ($h$) as input and retrieves the most relevant premises from the knowledgebase ($\mathbb{P}\subset\mathcal{K}$)~\cite{douze2024faiss}. The relevant premises are then augmented to the original input prompt and are used to generate rationales via the GPT-4.

\textbf{CILR} prompting employs a structured approach to guide a LLM’s question answering through provision of foundational context based on the premises retrieved from \textbf{CIKB}. The instructions to guide GPT-4's generative process included in the input prompt are as follows,
\textit{"You are an expert in judging a legal hypothesis on the criminal investigation domain. Given the hypothesis and premise, your task is to judge whether the legal hypothesis is True or False through the following process."}
\begin{itemize}
    \item \textit{First, you should analyze the given hypothesis based on the premise considering both the overall context and specific details.}
    \item \textit{Subsequently, if the premise provides sufficient information to judge the hypothesis, state your judgement based on the premise. If not, make a judgement of the legal hypothesis to the best of your legal expertise and reasoning abilities.}
    \item \textit{Use pertinent legal principles, precedents, and logical arguments to answer.}
\end{itemize}

To ensure that the SLM is trained on various legal reasoning approaches, we utilized six different prompting strategies for each data instance as follows,
\begin{itemize}
    \item \textbf{GPT4-VP-ZS}: Zero-shot vanilla prompting method.
    \item \textbf{GPT4-IRAC-ZS}: Zero-shot IRAC prompting method.
    \item \textbf{GPT4-IRAC-1S}: One-shot IRAC prompting method.
    \item \textbf{GPT4-CILR-ZS-CIKR}: Zero-shot CILR prompting for meth-od with CI Knowledge Retrieval.
    \item \textbf{GPT4-CILR-1S-CIKR}: One-shot CILR prompting for meth-od with CI Knowledge Retrieval.
    \item \textbf{GPT4-CILR-3S-CIKR}: Three-shot CILR prompting for meth-od with CI Knowledge Retrieval.
\end{itemize}

We used the GPT-4 to generate various responses for each data instance contained in our initial dataset $\mathcal{D}_i$. We then eliminated those that have incorrect assessment (i.e wrongly predicted true-false labels) of the given hypothesis. We then augmented our dataset with the GPT-4-generated rationales ($\mathcal{D}^{+}$) where each data instance is a pair of input prompt (crime investigation context $\mathcal{C}$, hypothesis $h$, premises $\mathbb{P}$) and output response (ground truth assessment $y_a$, GPT-generated rationale $y_r$). The training set ($\mathcal{D}^{+}_t$) contains 14,105 rationales for 3,204 hypotheses while the development set ($\mathcal{D}^{+}_d$) contains 465 rationales for 462 hypotheses.

\subsubsection{Domain Expert Curation}\hfill\\
Some GPT-generated rationales may contain legal grounds based on outdated criminal law articles or overruled court cases. Also, we found some hypothesis assessments containing a few number of rationales which may hinder the finetuning effects on the SLM. Moreover, there is a possibility that the GPT-4 model could generate misleading rationales (i.e hallcination) while predicting the correct true-false assessments for the given hypotheses. To cope with these issues, we collaborated with 20 law enforcement officials affiliated with the Korean National Police Agency to manually revise the GPT-generated rationales and annotate expert-level explanations for hypothesis assessment cases that lack sufficient rationale. As a result, our final version of the training set used in finetuning the SLM contains 14,105 data instances ($\mathcal{D}^{+}_t$ ($\mathcal{C}$, $h$, $\mathbb{P}$, $y_a$, $y_r$)) including those having rationales curated and annotated by domain experts. Table~\ref{tab:lapis_dataset} shows the detailed statistics of our finalized Crime Investigation Legal Reasoning dataset. 

\subsection{Finetuning Small Language Model}
We finetuned the SLM's Criminal Investigation Legal Reasoning task on the training set $\mathcal{D}^{+}_t$ while using the development set $\mathcal{D}^{+}_d$ for validation purposes. Given a data instance containing a pair of input prompt $\mathbf{X}$ (crime investigation context $\mathcal{C}$, hypothesis $h$) and output response $\mathbf{y}$ (ground truth True-False assessment $y_a$, supportive rationale $y_r$), the model parameters $\theta$ of the SLM $f$ optimized under loss objective $\mathcal{L}(\theta)$ so that it generates the desired response $\hat{\textbf{y}}$ that closely matches $\mathbf{y}$.

The model parameters ($\theta$) of the SLM optimized under loss objective $\mathcal{L}(\theta)$ are mathematically expressed as follows,
\begin{align}
\mathcal{L}(\theta) &= -\sum_{t=1}^{T}\mathbf{y}_t \log{(\hat{\mathbf{y}}_t)} \\
 &= -\sum_{t=1}^{T}\mathbf{y}_t \log{(f_{\theta}(\mathbf{X}))} \\
 &= -\sum_{t=1}^{T}\mathbf{y}_t \log{(f_{\theta}(\mathcal{C},h))} 
\end{align}

The SLM candidates whose model architecture and pretrained weights are imported from HuggingFace are the following~\cite{wolf2019huggingface},
\begin{itemize}
    \item \texttt{yanolja/EEVE-Korean-Instruct-10.8B-v1.0}
    \item \texttt{yanolja/EEVE-Korean-10.8B-v1.0}
    \item \texttt{yanolja/KoSOLAR-10.7B-v0.2}
    \item \texttt{beomi/Yi-Ko-6B}
    \item \texttt{beomi/llama-ko-7b}
    \item \texttt{beomi/OPEN-SOLAR-Ko-10.7B}
\end{itemize}

All SLM candidates were trained for 5 epochs using the same Paged AdamW optimizer with its learning rate set to 0.0001. We employed the Quantized version of Low Rank Adaptation (QLoRA)~\cite{dettmers2023qlora} for training efficiency. The hardware used for finetuning the SLMs is two NVIDIA A100 80GB Tensor Core GPU devices while the elapsed training time was approximately 30 hours.

\section{Experiments and Analysis}

\subsection{LAPIS Deployment}
LAPIS is a crime investigation system built based on two major components which are the evaluator and retriever. The evaluator $f$ is an SLM finetuned on our constructed dataset, is prompted with a question asking whether the user's formulated hypothesis $h$ based on current crime investigation context $\mathcal{C}$ is true or false and generates a response consisting of the true-false assessment $y_a$ and its supportive rationale $y_r$. Previously utilized during the rationale generation process, the retriever $g$ augments the input prompt with the top 5 relevant premises retrieved from the CIKB via vector similarity search $\mathbb{P}\subset\mathcal{K}$. In sum, LAPIS' overall generative process is mathematically expressed as $\mathbf{y} = f(h, \mathcal{C}, \mathbb{P})$ where $\mathbb{P} = g(h, \mathcal{C})$.

\subsection{Experimental Settings}

\begin{table*}[]
\scalebox{0.6}
{
\begin{tabular}{|l|l|l|cc|cc|cc|cc|}
\hline
\multicolumn{1}{|c|}{\multirow{2}{*}{\textbf{Model}}}                                                                          & \multicolumn{1}{c|}{\multirow{2}{*}{\textbf{Method}}} & \multicolumn{1}{c|}{\multirow{2}{*}{\textbf{CIKR}}} & \multicolumn{2}{c|}{\textbf{Criminal Law}}      & \multicolumn{2}{c|}{\textbf{Criminal Procedure Law}} & \multicolumn{2}{c|}{\textbf{Criminal Investigation}} & \multicolumn{2}{c|}{\textbf{Total}}             \\ \cline{4-11} 
\multicolumn{1}{|c|}{}                                                                                                         & \multicolumn{1}{c|}{}                                 & \multicolumn{1}{c|}{}                               & \multicolumn{1}{c|}{\textbf{ACC}} & \textbf{F1} & \multicolumn{1}{c|}{\textbf{ACC}}    & \textbf{F1}   & \multicolumn{1}{c|}{\textbf{ACC}}    & \textbf{F1}   & \multicolumn{1}{c|}{\textbf{ACC}} & \textbf{F1} \\ \hline
\texttt{EEVE-Korean-Instruct} (10.8B, ours)                                                                         & \textbf{CILR-FT}                                      & \multicolumn{1}{c|}{$\bigcirc$}                                          & \multicolumn{1}{c|}{0.78}         & 0.81        & \multicolumn{1}{c|}{0.75}            & 0.82          & \multicolumn{1}{c|}{\textbf{0.70}}            & \textbf{0.75}          & \multicolumn{1}{c|}{\textbf{0.74}}         & \textbf{0.79}        \\ \hline
\texttt{EEVE-Korean} (10.8B)                                                                                  & \textbf{CILR-FT}                                      & \multicolumn{1}{c|}{$\bigcirc$}                                          & \multicolumn{1}{c|}{0.72}         & 0.78        & \multicolumn{1}{c|}{\textbf{0.78}}            & \textbf{0.85}          & \multicolumn{1}{c|}{\underline{0.68}}            & \underline{0.75}          & \multicolumn{1}{c|}{0.72}         & 0.78        \\ \hline
\texttt{gpt4} (1.7T)                                                                                          & \textbf{CILR-3S}                                      & \multicolumn{1}{c|}{$\bigcirc$}                                          & \multicolumn{1}{c|}{\textbf{0.82}}         & \textbf{0.85}        & \multicolumn{1}{c|}{\underline{0.78}}            & \underline{0.84}          & \multicolumn{1}{c|}{0.65}            & 0.71          & \multicolumn{1}{c|}{\underline{0.73}}         & \underline{0.79}        \\ \hline
\texttt{gpt4} (1.7T)                                                                                          & \textbf{CILR-ZS}                                      & \multicolumn{1}{c|}{$\bigcirc$}                                          & \multicolumn{1}{c|}{\underline{0.79}}         & \underline{0.82}        & \multicolumn{1}{c|}{0.77}            & 0.84          & \multicolumn{1}{c|}{0.62}            & 0.69          & \multicolumn{1}{c|}{0.70}         & 0.76        \\ \hline
\texttt{OPEN-SOLAR-Ko} (10.7B)                                                                                & \textbf{CILR-FT}                                      & \multicolumn{1}{c|}{$\bigcirc$}                                          & \multicolumn{1}{c|}{0.75}         & 0.79        & \multicolumn{1}{c|}{0.72}            & 0.80          & \multicolumn{1}{c|}{0.64}            & 0.70          & \multicolumn{1}{c|}{0.69}         & 0.75        \\ \hline
\texttt{KoSOLAR} (10.7B)                                                                                      & \textbf{CILR-FT}                                      & \multicolumn{1}{c|}{$\bigcirc$}                                          & \multicolumn{1}{c|}{0.63}         & 0.72        & \multicolumn{1}{c|}{0.68}            & 0.79          & \multicolumn{1}{c|}{0.60}            & 0.70          & \multicolumn{1}{c|}{0.63}         & 0.73        \\ \hline
\texttt{LLaMA2-Ko} (7B)                                                                                       & \textbf{CILR-FT}                                      & \multicolumn{1}{c|}{$\bigcirc$}                                          & \multicolumn{1}{c|}{0.56}         & 0.70        & \multicolumn{1}{c|}{0.70}            & 0.82          & \multicolumn{1}{c|}{0.60}            & 0.74          & \multicolumn{1}{c|}{0.62}         & 0.73        \\ \hline
\texttt{Yi-Ko} (6B)                                                                                           & \textbf{CILR-FT}                                      & \multicolumn{1}{c|}{$\bigcirc$}                                          & \multicolumn{1}{c|}{0.51}         & 0.60        & \multicolumn{1}{c|}{0.59}            & 0.70          & \multicolumn{1}{c|}{0.42}            & 0.54          & \multicolumn{1}{c|}{0.49}         & 0.60        \\ \hline
\multirow{5}{*}{\begin{tabular}[c]{@{}l@{}}\texttt{EEVE-Korean-Instruct} \\ (Ablation Settings)\end{tabular}} & \textbf{CILR-FT (-DEXP)}                              & \multicolumn{1}{c|}{$\bigcirc$}                                          & \multicolumn{1}{c|}{0.75}         & 0.78        & \multicolumn{1}{c|}{0.71}            & 0.80          & \multicolumn{1}{c|}{0.51}            & 0.57          & \multicolumn{1}{c|}{0.63}         & 0.70        \\ \cline{2-11} 
                                                                                                                               & \textbf{CILR-FT}                                      & \multicolumn{1}{c|}{$\times$}                                            & \multicolumn{1}{c|}{0.66}         & 0.75        & \multicolumn{1}{c|}{0.74}            & 0.83          & \multicolumn{1}{c|}{0.61}            & 0.75          & \multicolumn{1}{c|}{0.66}         & 0.77        \\ \cline{2-11} 
                                                                                                                               & \textbf{CILR-ZS}                                      & \multicolumn{1}{c|}{$\times$}                                            & \multicolumn{1}{c|}{0.52}         & 0.45        & \multicolumn{1}{c|}{0.56}            & 0.61          & \multicolumn{1}{c|}{0.51}            & 0.50          & \multicolumn{1}{c|}{0.53}         & 0.52        \\ \cline{2-11} 
                                                                                                                               & \textbf{IRAC-ZS}                                      & \multicolumn{1}{c|}{$\times$}                                            & \multicolumn{1}{c|}{0.01}         & 0.02        & \multicolumn{1}{c|}{0.01}            & 0.02          & \multicolumn{1}{c|}{0.02}            & 0.04          & \multicolumn{1}{c|}{0.01}         & 0.03        \\ \cline{2-11} 
                                                                                                                               & \textbf{VP-ZS}                                        & \multicolumn{1}{c|}{$\times$}                                            & \multicolumn{1}{c|}{0.11}         & 0.17        & \multicolumn{1}{c|}{0.19}            & 0.30          & \multicolumn{1}{c|}{0.11}            & 0.14          & \multicolumn{1}{c|}{0.13}         & 0.20        \\ \hline
\end{tabular}
}
\caption{Evaluation results for each finetuned SLM's performance on its hypothesis assessment ability in Legal Reasoning Task. \textbf{CILR-FT} refers to finetuning the SLM with data using Crime Investigation Legal Reasoning prompting method. \textbf{CILR-3S} and \textbf{CILR-ZS} refers to prompting the pretrained model using 3-shot and zero-shot learning respectively. \textbf{IRAC-ZS} and \textbf{VP-ZS} refers to prompting the model with IRAC and Vanilla prompting method respecively. \textbf{-DEXP} refers to removing data instances containing expert-annotated rationales. Bold-faced and underlined scores refer to best and second-best performance respectively. \textbf{Total} refers to the scores calculated for all questions across the three subjects.} 
\label{tab:lapis_results}
\end{table*}

Upon selecting the most suitable SLM to be used as LAPIS' evaluator, we first performed experiments with different SLM candidates and also included GPT-4 (\texttt{gpt-4-1106-preview}) as our baseline. We finetuned all pretrained SLMs with the same optimization setting and prompting method (\textbf{CILR-FT}). We then ran LAPIS' model inference on the test partition of the dataset ($\mathcal{D}_s$). 

Due to the difficulty of finetuning GPT-4, we instead used its API to generate the responses for the same test set, using the same \textbf{CILR} prompting and \textbf{CIKR} method. The GPT-4 baseline experiments were conducted with zero-shot and 3-shot learning settings (\textbf{CILR-ZS} and \textbf{CILR-3S}).

All model setting's performance were evaluated using the classif-ication-related metrics, accuracy (ACC), F1-score (F1) based on their generated binary-labeled assessments (True or False) compared with ground truths ($y_a$) per data instance. Table~\ref{tab:lapis_results} shows each setting's performance on legal hypothesis assessment. 

According to the first eight rows in Table~\ref{tab:lapis_results}, finetuning the pretrained weights of \texttt{EEVE-Korean-Instruct} model showed the best hypothesis assessment performance in terms of accuracy and F1-score (0.74, 0.79).  Interestingly, \texttt{EEVE-Korean-Instruct} demonstrated its legal reasoning ability quantitatively sub-par with GPT4 (0.73, 0.79), despite having approximately 1/157 the number of model parameters. This suggests that not only choosing the most adequate SLM but also incorporation of domain-specific approaches such as retrieving the relevant premises and finetuning with expert-curated rationales could potentially enhance LAPIS' ability to guide police's investigative decisions. Also, we emphasize that our system design approach minimizes reliance on proprietary LLMs such as GPT-4. 

We performed ablation studies by removing LAPIS' Crime Investigation Knowledge Retrieval capability (\textbf{CILR-FT} without \textbf{CIKR}), eliminating data instances containing expert-annotated rationales (\textbf{CILR-FT (-DEXP)}, skipping the finetuning phase (\textbf{CILR-ZS}) and applying other prompting methods (\textbf{IRAC-ZS}, \textbf{VP-ZS}). 

Overall, the ablation results show that removing the expert-annotated rationales from the training set used for finetuning the SLM, incurs performance loss in providing the correct assessment for given hypotheses (0.63, 0.70). Also, prompting language models without reliance on \textbf{CIKR}~is detrimental (0.66, 0.67), as the premises extracted from the \textbf{CIKB} play a key role in helping them make accurate legal assessments on hypotheses. Notably, other prompting methods especially the IRAC prompting method (0.01, 0.03) fell extremely behind our proposed CILR prompting method even without finetuning (0.53, 0.52) or vanilla prompting method (0.13, 0.20).

\section{Analysis}

\begin{figure}[h]
    \scalebox{0.7}
    {
    \centering
    \includegraphics[width=0.49\textwidth]{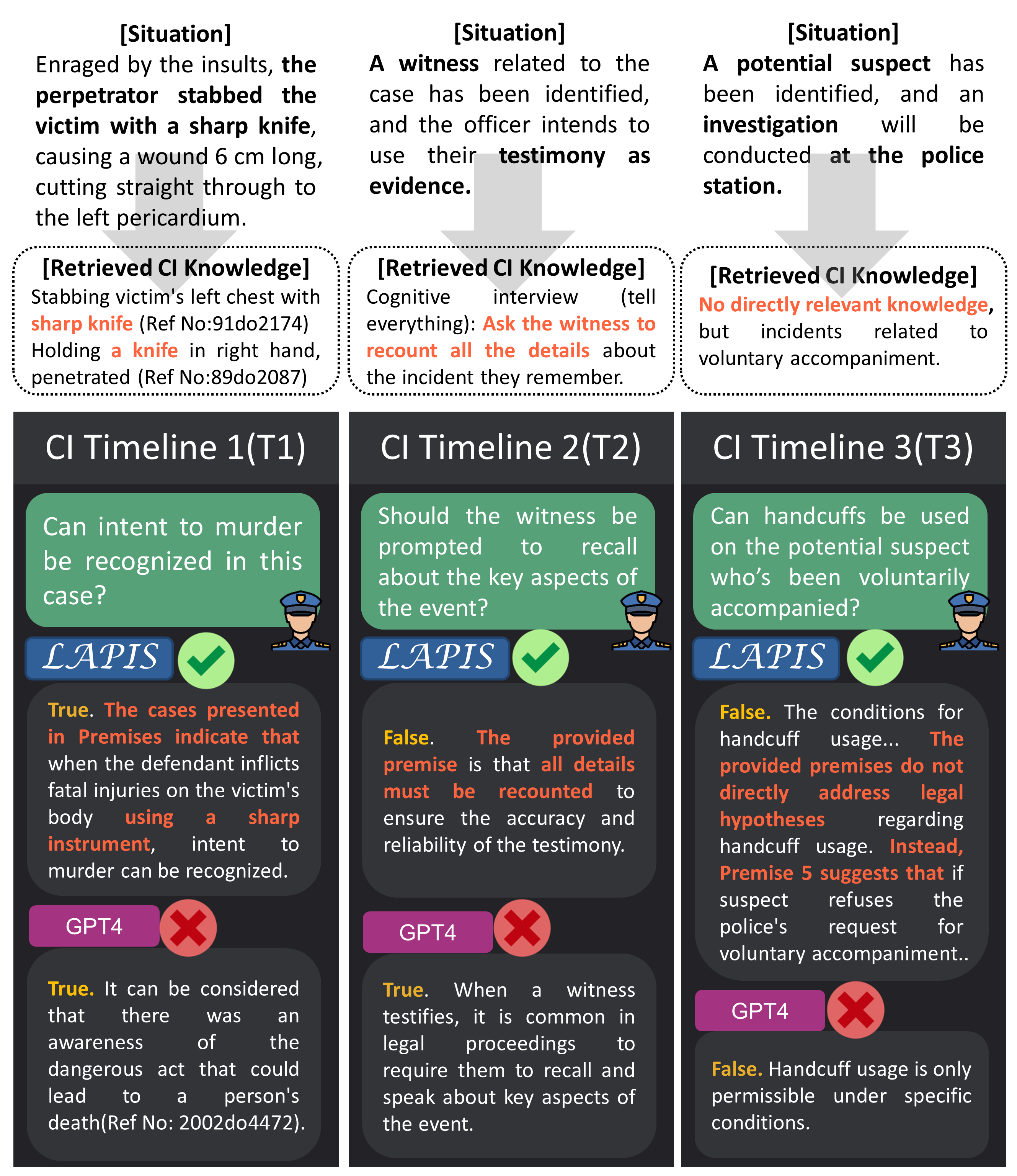}
    }
	\caption{Utilization of LAPIS responses assuming a murder case. LAPIS appropriately references the provided \textbf{CI Knowledge} and assists in the process of criminal investigation.
	}
	\label{fig:realcase}
\end{figure}

We deployed LAPIS in an actual crime investigation scenario and evaluated its practicability based on its generated response in comparison with GPT-4. Throughout the timeline of this scenario, the current context of crime investigation changes, bringing new legal hypotheses to the police officer regarded as the end user of this system. Figure~\ref{fig:realcase} illustrates how LAPIS offers legal guidance to the police officer at each step of this timeline through accurate assessment of their hypothesis associated with its generated rationale specialized in crime investigation knowledge.

\subsection{(T1) Initial Assessment of Crime Scene}
At this step, the police officer receives an initial witness report about the victim being stabbed by the perpetrator's knife which caused excessive bleeding that eventually lead to victim's death. The autopsy report released by the National Forensic Service states that the wound was 6cm long, 17cm deep, cutting through the left pericardium. Having known that this incident began with a quarrel between the victim and perpetrator, the police officer questions a legal hypothesis of whether murder intent can be recognized in this case.

The retrieved premises from the crime investigation knowledgebase are the following,
\begin{itemize}
    \item Given that the accused stabbed the victim's heart with a fruit knife held in their right hand and fled immediately after the victim died from excessive blood loss, the depth and location of the wound suggest that this was not an accidental action during the fight but rather an intentional act, carried out with awareness of the potential for murder (Ref No: 89do2087).
    \item Given that the accused stabbed the victim's hypogastrium with a kitchen knife, resulting in a wound that was 5 cm deep and 15 cm long, leading to the victim's death due to excessive bleeding, it is suggested that even if the accused did not have a strong intent to commit murder, they may have exhibited gross negligence. The consequences of such a stabbing could have been reasonably predicted (Ref No:82do2525).
    \item If the accused stabbed the victim's left armpit with a sashimi knife, directing the blow towards the heart and causing the victim to die within an hour, it is plausible to assume that, even if this act happened suddenly, the accused was aware that their action could result in manslaughter (Ref No:83do1594).
\end{itemize}

By appropriately exploiting the retrieved court ruling precedents utilized as premises, LAPIS correctly assessed the police officer's hypothesis as correct, leading them to narrow down the nature of this case into intended murder and proceed to the next step which is identification of the perpetrator by interacting with witnesses. On the contrary, while GPT4 made the correct hypothesis assessment, its generated ratioanle contained a non-existent precedent.

\subsection{(T2) Acquisition of Witness Statements}
At this step, the police officer has recognized the essence of this criminal case which is intended murder. A witness contacts the police officer, claiming that they can describe the perpetrator's appearance which may lead to their identification. The police officer thinks that obtaining only the key details may not only help identify the perpetrator but also bring forward closing the case. This hypothesis should be deemed as False since the implied investigative action may lead to cognitive investigation bias.

The retrieved premises from the crime investigation knowledgebase are the following,
\begin{itemize}
    \item The witness should disclose everything, even details that seem minor or irrelevant to the criminal case. Every single piece of information can be gathered to potentially become significant clues in this investigation (textbook).
    \item The investigator should encourage the witness to recall their psychological and physical state during the incident. Additionally, the investigator should guide the witness to remember the details step by step, gradually reconstructing the entire context. Allowing intervals of a few seconds between each step can help the witness recall their memory with greater precision and detail.
    \item A person's memory based on their experience may fade over time rather than the opposite. Therefore, if the testimony of a witness repeatedly changes over time towards alignment to the fact of indictment without any specific reason, its credibility must be questioned (Ref No:84do22).
\end{itemize}

LAPIS successfully assessed the hypothesis as False, stating one of the fundamental principles stated in crime investigation textbooks which is "testimony should be obtained as detailed as possible". However, GPT4 solely relied on its parametric knowledge, producing incorrect information that did not effectively prompt detailed witness testimony.

\subsection{(T3) Prevention of Illegal Coercive Measures}
At this step, the suspect identified by the police officer volunteering came to the police station in order to take part in the investigation. However, as the suspect is posed with provocative questions by the police officer, they refuse to answer any more questions and decides to leave the police station. The police officer considers to prevent the suspect from leaving and formulates an hypothesis whether it is legal to perform coercive measures.

In this case, there are no premises directly relevant to the abovementioned crime investigation context and hypothesis. Despite that, LAPIS does not fully exploit the premises but extrapolates its legal reasoning process based on one of the retrieved premises which is a precedent that concludes the eligibility of using coercive measures only when the suspect is exhibiting violent behavior. On the other hand, GPT-4 fails to cite appropriate grounds and provides less specific responses, highlighting a difference in practicability.

\section{Conclusion and Future Work}
We introduce LAPIS, a police investigation system consisting of two main components which are the evaluator and retriever. The evaluator is a finetuned SLM which showed notable performance in the legal reasoning task. The retriever provides the evaluator with relvant premises automatically obtained from our created domain-specific knowledgebase. In addition, simulation results on LAPIS being deployed in a real-world crime scene support our pipeline design. As the central approach of our work is enhancing LM with criminal investigation specialized legal reasoning skills, we expect it to be one of key insights in developing AI-assisted crime investigation systems. 

However, there are several limitations of this work to be addressed. First, our constructed dataset, knowledgebase and finetuned SLM is exclusively based on Korean language. We plan to seek ways to incorporate multilingual modeling approaches so that other nations can benefit greatly from our system design approach. Second, even though we used 10 years of released national exam questions to construct our dataset, the lack of training data instances address the need of data augmentation techniques. We plan to employ automated approaches to generate more training data built from recently updated criminal cases. Finally, while LAPIS exhibits promising performance in legal reasoning tasks, it does not explicitly suggest possible investigative actions to the police officer. We plan to add this functionality to facilitate more practicability for law enforcement agencies.

Our work is part of the development project affiliated Korean National Police Agency. The aim of this study is to develop an AI-assisted crime investigation decision support system tailored for police officers. We expect LAPIS and its future work to benefit criminal investigators in not only Korea but also other countries and result in reduction of crime rates.\\\\

\begin{acks}
This work was supported in part by the National Research Foundation of Korea (NRF-2023R1A2C3004176), the MSIT (Ministry of Science and ICT), Korea, under the ICT Creative Consilience program (IITP-2023-2020-0-01819) supervised by the IITP (Institute for Information \& communications Technology Planning \& Evaluation), the MCST (Ministry of Culture, Sports and Tourism), Korea, under the Cultural Service Expansion Technology Development Program, 

This work was supported by Institute of Information \& communications Technology Planning \& Evaluation (IITP) grant funded by the Korea government(MSIT) (IITP-2022-0-00653, Voice Phishing Information Collection and Processing and Development of a Big Data Investigation Support System)

Donghee Choi is additionally supported by the Horizon Europe project CoDiet. The CoDiet project is funded by the European Union under Horizon Europe grant number 101084642 and UK Research and Innovation (UKRI) under the UK government's Horizon Europe funding guarantee.
\end{acks}

\bibliographystyle{ACM-Reference-Format}
\balance
\bibliography{9_reference.bib}

\end{document}